\documentclass[12pt]{l4dc2021} 

\usepackage{multicol}
\usepackage{comment}
\newcommand{\E}{\mathbb{E}}
\newcommand{\Prob}{\mathbb{P}}

\usepackage{dsfont}
\newcommand{\ind}{\mathds{1}}
\title[The Dynamics of Gradient Descent for Overparametrized Neural Networks]{The Dynamics of Gradient Descent for Overparametrized Neural Networks}
\usepackage{times}

\author{%
 \Name{Siddhartha Satpathi} \Email{ssatpth2@illinois.edu}
 \AND
 \Name{R Srikant} \Email{rsrikant@illinois.edu}\\
 \addr Department of Electrical and Computer Engineering, University of Illinois at Urbana Champaign%
}

\begin{document}
\maketitle

\begin{abstract}%
 We consider the dynamics of gradient descent (GD) in overparameterized single hidden layer neural networks with a squared loss function. Recently, it has been shown that, under some conditions, the parameter values obtained using GD achieve zero training error and generalize well if the initial conditions are chosen appropriately. Here, through a Lyapunov analysis, we show that the dynamics of neural network weights under GD converge to a point which is close to the minimum norm solution subject to the condition that there is no training error when using the linear approximation to the neural network. To illustrate the application of this result, we show that the GD converges to a prediction function that generalizes well, thereby providing an alternative proof of the generalization results in \cite{arora2019fine}.
\end{abstract}


\section{Introduction}
Neural Networks have shown promise in many supervised  learning tasks like image classification \cite{ImageNet} and semi-supervised learning like reinforcement learning \cite{DQNatari}. A key reason for the success of neural networks is that they can approximate any continuous function with arbitrary accuracy \cite{hornik,cybenko}. Further choosing the neural network parameters by optimizing a loss function over this complex non-convex function class using simple first order methods, like gradient descent, leads to good generalization for unseen data points. We address the optimization and generalization aspect of neural network training in this paper. Recently it has been shown that when the network is overparametrized, i.e, the number of neurons in a layer is much larger than the number of training points, one can achieve zero training loss by running gradient descent on the squared loss function \cite{du18}. This line of research is motivated by the fact when one runs gradient descent on an overparametrized network initialized appropriately, the network behaves as though it is linear function of its weights \cite{jacot2018neural,lee2019wide}. Although one can achieve zero loss in an overparametrized network, there may be more than one set of network weights which achieve zero loss.
In this paper, we study the dynamics of gradient descent for a single hidden layer, overparameterized neural network, and show that gradient descent converges to a certain minimum norm solution. We present an application of such a characterization of the limit behavior of the network weights by providing an alternative proof of a generalization result in \cite{arora2019fine}.

We are motivated by the results in \cite{du18} where it was shown that if the network width is polynomial in the number of data points, gradient descent converges to the global optimum of the squared loss function. The main idea behind their proof is to show that an overparametrized neural network behaves similarly to a linear function (linear in the weights). 
The linear approximation can be described as follows: the input is mapped to a high-dimensional feature vector and the linear approximation is an inner product of the network weights and this high-dimensional feature vector. In the limit as the number of neurons goes to infinity, this mapping of the input to a high-dimensional space can also be viewed in terms of a kernel corresponding to a reproducing kernel Hilbert space (RKHS). Such a kernel is known as the Neural Tangent Kernel (NTK) and was introduced in \cite{jacot2018neural}. In this paper, we prove that the weights of an overparametrized neural network converge close to the point where the weights of the linearized neural network would have converged had we used the linearized network in the optimization process. To the best of our knowledge, the results in earlier papers do not address such convergence properties.

\subsection{Related work}
\begin{itemize}
    \item \textbf{Convergence of Gradient Descent for squared loss:} For the squared loss function and ReLU activation function, the convergence of gradient descent was proved in \cite{du18}.  This is further extended to deep networks in \cite{du2019,allen2018learning}, but the dependence of width grows exponentially with the depth of the network; see \cite{lee2019wide,chizat2019lazy} for analysis in case of general differentiable activation functions. The dependence of the width of the network in terms of the data points has been further improved in \cite{oymak2020} by careful choice of Lyapunov functions.
    \item \textbf{Convergence of Gradient Descent for logistic loss:}
    Overparametrized neural network with logistic loss function were shown to be in the linear regime for a finite time in \cite{ji2019}. Under assumptions that the data distribution is linearly separable by the neural tangent kernel, it is shown that gradient descent reaches good test accuracy in this finite time. Unlike the squared loss, optimizing the logistic loss only requires a poly-logarithmic dependence on the number of data points. This is further extended to deep Networks in \cite{chen2019deep}.
    \item \textbf{Implicit Regularization:}
    This line of work is closest to our work (see \cite{neyshabur2017implicit} for an overview). It is known that for least squares linear regression, the iterates of gradient descent converge to the minimum norm solution subject to zero loss. This is also true for certain nonlinear models as shown in \cite{oymak2019overparameterized}. We further add to this literature by showing that the iterates of the neural network weights under gradient descent stay close to the minimum norm solution of the linearized model and this distance decreases with the increase in the width of the network. Since gradient descent chooses a particular characterization of weights from all possible solutions achieving zero training error, this phenomenon is often referred to as the `implicit bias' or `implicit regularization.' For logistic loss, the weights diverge to infinity, but implicit regularization still occurs. Specifically, with linear classification, the direction along which the weights diverge to infinity matches the direction of the hard margin SVM solution when the data is separable \cite{ji2018risk,soudry2018implicit}. This result has been extended to some classes of neural networks recently in \cite{chizat2020implicit}. 
    \item {\textbf{Generalization results for squared loss: } Shallow Neural Networks are shown to generalize well for a large class of functions in \cite{arora2019fine}. We provide an alternative proof of their result for the case of the squared loss function. In \cite{montanari}, th authors consider the generalization properties of Kernel Ridge Regression for NTK under different activation functions, but their results are not directly applicable to finite-width networks as is the case in this work.} 
\end{itemize}

\section{Problem Statement and Contribution}
Consider a neural network which takes $x$ as its input and produces an output $f(x, w,b)$ where $w,b$ are certain weight parameters to be chosen. We suppose that $f:\mathbb{R}^d\times\mathbb{R}^{md}\times \mathbb{R}^{m}\rightarrow \mathbb{R}$ is of the form
\begin{align*}
    f(x,w,b) := \sum_{k=1}^m \frac{a_k}{\sqrt{m}} \sigma(w_k^Tx+b_k)
\end{align*}
where $\sigma(.) = \max(0,.)$ is the Rectified Linear Unit (ReLU) activation function and $x\in\mathbb{R}^d.$ This describes a neural network with one hidden layer where the input weights are denoted by $w$'s, input biases are $b$'s and output layer weights are $a$'s. We absorb the biases $b$'s as an extra dimension in the weight vector $w$'s. Likewise, let $\tilde{x} = \{x,1\}.$ So we can compactly write the neural network function as $f: \mathbb{R}^d\times\mathbb{R}^{m(d+1)}\rightarrow \mathbb{R}.$
\begin{align*}
    f(x,w) := \sum_{k=1}^m \frac{a_k}{\sqrt{m}} \sigma(w_k^T\tilde{x})
\end{align*} 

We are given $n$ data points $\{x_i,y_i\}_{i=1}^n$ drawn i.i.d. from a distribution $\mathcal{X}\times\mathcal{Y}$ and we would like to minimize the mean-square error  $$L(w)=\sum_{i=1}^n (y_i - f(x_i,w))^2$$
over all $w.$ One of the ways to perform the above minimization is to use the Gradient Descent (GD) algorithm. GD is an iterative algorithm where in each step $w$ is updated in the direction of the negative gradient of $L(w).$ Given appropriate initialization of $w(0),$ and step size $\eta$ for $k=1,2,\dots$

$$
w(k+1) = w(k) - \eta \frac{\partial}{\partial w} L(w)\Big{|}_{w = w(k)}
$$
GD is known to converge to a global optimum if $L(w)$ were a convex function of $w$ for small enough values of $\eta.$ In our case, even if $L(w)$ is non convex, \cite{du18} show that GD initialized appropriately converges to a global optimum, $i.e. L(w(k))$ goes to $0$ as $k\rightarrow \infty$. 

We would like to characterize the performance of $w(k)$ given a new data point sampled from the same distribution $\mathcal{X}\times\mathcal{Y}$. In particular, we are interested in the  generalization error for $w(k)$ which is defined as 

$$
E_{x\times y\sim\mathcal{X}\times\mathcal{Y}} (y-f(x,w(k)))^2.
$$
We assume that the samples of $x$ are chosen from $\mathcal{X}$. Given $x,$ $y$ is conditionally chosen as $y = f^*(x) + \zeta,$ where $\zeta$ is drawn from a mean zero distribution independent of $\mathcal{X}$ and $f^*$ is a continuous function to be specified later.


For the noise less case, i.e. $\zeta = 0,$ the generalization error has been characterized in \cite{arora2019fine} for certain choices of $f^*.$ For the noisy case, to the best of our knowledge the generalization error has not been characterized. But, \cite{montanari} consider the generalization error for the linearized version of the neural network.
We first explain this linearization.

Consider the first order Taylor approximation of $f(x,w)$ around $w(0).$ Define
$$
f_L(x,w) := f(x,w(0)) + \nabla^{\top} f(x,w(0)) (w - w(0)) = \frac{1}{\sqrt{m}}\sum_{k=0}^m a_k\ind_{w_k^{\top}(0)\tilde{x}\ge 0}w_k^{\top}\tilde{x}.
$$
It was shown in \cite{du18} that $w$ stays close to $w(0)$ during the training iterations of gradient descent when then neural network is overparametrized, $i.e., m>poly(n).$ As a result, $f(x_i,w)$ is close to $f_L(x_i,w)$ for $i\in [n]$ during the GD iterations. This makes us consider the following problem. Instead of minimizing $L(w),$ consider the loss function $L_{lin}(w) = \sum_i (y_i-f_L(x_i,w))^2.$ Minimizing $L_{lin}(w)$ using GD is expected to achieve zero loss since  $L_{lin}(w)$ is a convex function. Since the network is overparametrized, there are inifinitely many values of $w$ achieving zero loss. However, it is known that GD finds the solution which has the lowest $\ell_2$ norm. So, minimizing $L_{lin}(w)$ using gradient descent would lead to a solution

\begin{align}
    w_L^*: = \arg\min_w \|w-w(0)\| \;\;s.t.\;\;f_L(x_i,w) = y_i, i = \{1,2,\ldots,n\}.\label{eq:minnorm}
\end{align}

For appropriately chosen $w(0),$ the prediction function $f_L(x,w_L^*)$ can be shown to be close to $\sum_{i=1}^n c_i K^{(m)}(x,x_i),$ for some constants $c_i$ and the kernel function $$K^{(m)}(x,x_i) = \frac{1}{m}\sum_{k=1}^m x^{\top}x_i\ind_{w_k^{\top}(0)x\ge 0}\ind_{w_k^{\top}(0)x_i\ge 0}.$$ In the infinite width limit, $i.e., m\rightarrow\infty,$ the kernel $K^{(m)}$ converges to a kernel $K$ which is independent of the initialization $w(0)$ \cite{jacot2018neural}.
If $K(x,y) = \phi^{\top}(x)\phi(y),$ then by Moore-Aronszajn theorem, the
RKHS $\mathcal{H}$ induced by $K$ is given by this: A function $g\in\mathcal{H}$ can be defined by $g(.)=\phi^{\top}(.)w_g$ and inner product between functions $f$ and $g$ in $\mathcal{H}$ is given by $\langle f,g\rangle_{\mathcal{H}}=w_g^{\top}w_f.$ Further, by Representer theorem, the solution to kernel regression in $\mathcal{H}$ would be given by
\begin{align}
    f_{KR}(.) &= \min_{g\in\mathcal{H}} \|g\|_{\mathcal{H}} \text{ s.t. } y_i = g(x_i)\nonumber \\
    &=\sum_{i}^n c_iK(.,x_i) \;\;\text{ s.t. }\;\; \sum_{i=1}^n c_iK(x_j,x_i)=y_j \text{ for } j=\{1,2,\ldots,n\}.\label{eq:KR}
\end{align}
Hence, in the infinite width limit, the prediction function $f_L(x,w_L^*)$ is close to the solution of kernel regression on the RKHS induced by $K.$ 
Now we discuss our contributions below.
\begin{itemize}
    \item The results in \cite{montanari} analyse the generalization properties of $f_{KR}$ in the presence of noise. They would apply to our paper in the following manner. 
\begin{align}
E_{x\sim\mathcal{X},\zeta} (y-f(x,w(k)))^2&\le 2E_{x\sim\mathcal{X},\zeta} (y-f_{KR}(x))^2 + 2E_{x\sim\mathcal{X}} (f(x,w(k))-f_{KR}(x))^2. \label{eq:genM}
\end{align}
If we show that the second term in the RHS of \eqref{eq:genM} is small, then the generalization results in \cite{montanari} applies to our setting. Our goal is to show that the second term is small and in order to do so we require to show a relation between $w(k)$ and $w_L^*$. Informally, we show that
\begin{align}
    \|w(k) - w_L^* \| = O\left(\frac{1}{m^{0.125}}\right) \text{ for } m \ge poly(n) \text{ w.h.p for large enough } k. \label{eq:mainres}
\end{align}
It was observed that  in \cite{du18,arora2019fine} that the weights stay in a ball around initialization, i.e., $\|w(k)-w(0)\|=O(poly(n))$.
Equation \eqref{eq:mainres} is a more finer characterization of $w(k)$ than \cite{du18}.

\item In the noiseless case, \cite{arora2019fine} provided generalization bounds for the prediction function $f(x,w(k)).$ We provide an alternate derivation motivated by the results in \cite{bartlett2020benign} in the noiseless case instead of the Rademacher complexity approach in \cite{du2019}. The results in \cite{bartlett2020benign} are derived from the point of view of linear regression. Thus it naturally applies to providing bounds to kernel regression in a RKHS. In the noise-free setting, ($\zeta =0$) we derive a bound on the first term in \eqref{eq:genM}, i.e, $E_{x\sim\mathcal{X}} (y-f_{KR}(x))^2.$
\end{itemize}

The structure of the paper is as follows. We establish the result \eqref{eq:mainres} for a continuous-time version of gradient descent in section \ref{sec:CGD}. This develops an intuitive understanding of the proof. In section \ref{sec:GD} this is extended to the discrete-time GD. In section \ref{sec:gen}, we use the results  developed on the convergence of weights under GD in section \ref{sec:CGD}-\ref{sec:GD}, to show that $E_{x\sim\mathcal{X}} (f(x,w(k))-f_{KR}(x))^2$ is small. In addition, $E_{x\sim\mathcal{X}} (y-f(x,w(k)))^2$ is shown to be small for the noise-free case.

\subsection{Notation}
Let $w(0)$ denote the point at which gradient descent is initialized. The observation vector is $Y = [y_1\;\;y_2\ldots y_n]^T$ and the neural network function for the inputs $\{x_i\}_{i=1}^n$ is $f = [f(x_1,w)\ldots f(x_n,w)]^T.$ At initialization $f_0 = [f(x_1,w(0))\ldots f(x_n,w(0))]^T.$ Let the ReLU activation function be $\sigma(x) = \max(x,0).$ The gradient of $f(x,w)$ w.r.t $w$ is $\nabla f(x,w) = [\frac{a_1}{\sqrt{m}}\sigma'(w_1^T\tilde{x})\tilde{x}^T\ldots \frac{a_m}{\sqrt{m}}\sigma'(w_m^T\tilde{x})\tilde{x}^T]^T.$ Denote the $m(d+1)\times n$ matrix $\nabla f = [\nabla f(x_1,w)\ldots  \nabla f(x_n,w)]$ and the gradient matrix at initialization
$\nabla f_0  = [\nabla f(x_1,w(0))\ldots  \nabla f(x_n,w(0))]$. Also $H = \E_{w(0)} \left(\nabla^T f_0 \nabla f_0\right).$ Let the projection matrix onto the column space of $\nabla f_0$ be $P_0= \nabla f_0 (\nabla^T f_0 \nabla f_0)^{-1}\nabla^T f_0$ and the projection matrix onto the null space be $P_0^{\perp} := I - P_0.$ When we write $\|.\|$ it means the $2$ norm if $.$ is a vector or the operator norm if $.$ is a matrix.
\subsection{Assumptions}
\label{sec:assump}
\begin{itemize}
    \item The data points are bounded, i.e., $|y_i|\le C_y,\|x_i\|= \sqrt{d},\forall i=1,\ldots,n.$
    \item No two data points $x_i,x_j$ are parallel to each other, $x_i\nparallel x_j.$   
\end{itemize}

The assumptions stated above are mild and the same as in the literature \cite{du18}. The second assumption essentially ensures that the matrix $H = \E_{w(0)}[\nabla^T f_0 \nabla f_0]$ is positive definite \cite[Theorem 3.1]{du18}. We reprove this fact in the Lemma below. Note that $H$ does not depend on the width of the network. 
\begin{lemma}
Define $\theta_{min}$ by $\cos\theta_{\min} := \frac{\max_{i\neq j}x_i^Tx_j + 1}{d+1}.$ Under the above assumptions, the smallest eigenvalue of $H = \E_{w(0)}[\nabla^T f_0 \nabla f_0]$ is strictly  positive, $H\succeq cI,c>0$ and 
\begin{align*}
    \Omega\left(\frac{(d+1)\theta_{\min}}{\sqrt{\log(2n)+1}}\right) =c\le \lambda_{\min}(H) =O((d+1)\theta_{\min})
\end{align*}
 when $\theta_{\min}<1.$ More complicated expressions for the lower and upper bounds which do not require the condition $\theta_{\min} <1$ can be found in the proof of this lemma in the appendix.
 \label{lem:psd}
\end{lemma}

    

\section{Continuous-Time Gradient Descent Algorithm}
\label{sec:CGD}
First we describe the continuous time gradient descent algorithm below.
\begin{itemize}
    \item Initialize $w_k(0) \sim \mathcal{N}(0,\kappa^2 I_d)$. $a_k$'s are initialized as $1$ with probability $1/2$ and $-1$ with probability $1/2$.
    \item Run gradient descent in continuous time, $\dot{w} = \frac{\partial L(w)}{\partial w} := -\nabla f(f-Y)$
\end{itemize}

\subsection{Training loss and bound on $\|w_k - w_k(0)\|$}
\label{sec:w_k}

The training loss goes to zero exponentially fast. This is shown in Theorem 3.2, \cite{du18}. 

\begin{lemma}[\cite{du18}]
\label{lem:trainloss}
The continuous-time gradient descent algorithm achieves zero loss with probability greater than $1-\delta$ (where the randomness is due to the initialization) when $m= \Omega\left(\frac{n^6d^4C_y^2}{c^4\delta^3}\right)$ and $\kappa=1.$ Further, the rate of convergence can be characterized as
\begin{equation}
    \|f(t)-Y\|\le \exp(-ct/4)\|f_0-Y\|.
\end{equation}
Moreover, the weights $w_k$'s remain in a small ball around the initialization $w_k(0)$ in the following sense:
\begin{equation}
    \|w_k(t)-w_k(0)\|= O\left(\frac{\sqrt{dn}}{c\sqrt{m}}\right)\|f_0-Y\|.
\end{equation}
\end{lemma}


\subsection{Bound on $\|w-w_L^*\|$}
We can show that $w_L^*$ from \eqref{eq:minnorm} is equal to
\begin{align*}
    w_L^* = P_0^{\perp}w(0) + \nabla f_0 (\nabla^T f_0\nabla f_0)^{-1}Y
\end{align*}
since $\nabla^T f_0\nabla f_0$ would be a positive definite matrix.
We prove that that the weights $w(t)$ converge to a point close to $w_L^*$ and the distance decreases when the number of neurons $m$ increases.

\begin{theorem}
If the number of neurons $m=\Omega\left(\frac{n^6d^4C_y^2}{c^4\delta^3}\right) ,$ then under assumptions from section \ref{sec:assump}, with probability greater than $1-\delta$ over initialization and $\kappa=1$
\begin{align*}
    \|w(t) - w_L^*\| \le \exp(-c/2t) \|w(0) - w_L^*\| + O\left(\frac{(dn)^{2.5}C_y^{1.5}}{c^{2.5}\delta^{1.5}m^{0.25}}\right).
\end{align*}
\label{thm:cont}
\end{theorem}
\textbf{Proof Idea:} Consider the dynamics of $w$, $$\dot{w} = \frac{\partial L(w)}{\partial w} := -\nabla f(f-Y).$$ Since $w(t)$ is close to $w(0)$ from Lemma \ref{lem:trainloss}, we can show that $\nabla f\approx \nabla f_0$ throughout the dynamics of $w.$ Hence $\dot{w}$ approximately lies in the column space of $\nabla f_0$. So $P_0^{\perp}\dot{w}$ can be expected to be small. To capture this intuition we choose the Lyapunov functions
\begin{align}
V_{\perp} := \|P_0^{\perp}(w - w_L^*)\|^2 \text{ and } V_{\parallel} := \|P_0(w - w_L^*)\|^2.\label{eq:lya} 
\end{align}
The proof of this theorem  is deferred to the Appendix \ref{app:thmcont}.

\section{Discrete-Time Gradient Descent algorithm}
\label{sec:GD}
In this section we present the discrete-time gradient descent algorithm, and show results similar to the continuous-time version.
\begin{itemize}
    \item Initialize $w_k(0) \sim \mathcal{N}(0,\kappa^2 I_d)$. $a_k$'s are chosen to be $1$ with probability $1/2$ and $-1$ with probability $1/2$. Choose a step size $\eta>0.$
    \item Run gradient descent, i.e., for $k=0,1,\ldots$, update the weights as
    
    $$w_{k+1} = w_k - \eta \nabla f (f- Y).$$
\end{itemize}

\subsection{Bound on $\|w-w_L^*\|$ for Gradient Descent}
The analysis of the gradient descent in discrete time is similar in spirit to that in continuous time. Hence we relegate the proof of Theorem \ref{th:GD} to the arXiv report.
Unlike in continuous time, Theorem \ref{th:GD} requires more neurons for the result to hold, $O\left(\frac{1}{m^{1/8}}\right)$ as opposed to $O\left(\frac{1}{m^{1/4}}\right)$.
\begin{theorem}
If the number of neurons $m=\Omega\left(\frac{(dn)^{10}C_y^6}{c^{10}\delta^6}\right), \kappa = 1 $ and $\eta = O\left(\frac{c}{(dn)^2}\right)$, then under assumptions from section \ref{sec:assump}, with probability greater than $1-\delta$ over initialization 
\begin{align*}
    \|w(k) - w_L^*\| \le \left(1 - \frac{c\eta}{2} \right)^{k/2}\|w(0)-w_L^*\| + O\left(\frac{(dnC_y)^{1.5}}{c^{1.5}\delta^{1.5} m^{0.125}}\right)
\end{align*}
for $k=0,1,\ldots,$. Also, $w(k)$ converges to some  point $w^*$ as $k\rightarrow\infty.$
\label{th:GD}
\end{theorem}

\section{Generalization}
\label{sec:gen}
{In this section we provide generalization results for the output of gradient descent. The analysis depends on the results from Theorem \ref{th:GD} from optimization. 
Suppose the data points $\{x_i\}_{i=1}^n$ are sampled from a distribution $\mathcal{X}$. In Lemma \ref{lem:gen} below we show that the prediction function $f(x,w(k))$ at the end of iteration $k$ of gradient descent is close to the minimum norm interpolator for Kernel Regression from \eqref{eq:KR}. We can show that the solution to \eqref{eq:KR} is given by $f_{KR}(x)$ given below. 

\begin{align}
    f_{KR}(x) &= h^T(x)H^{-1}Y,\; h(x) :=[K(x,x_i),i=1\text{ to } n],\\
    H_{ij} &= K(x_i,x_j) = \tilde{x}_i^T\tilde{x}_j\frac{\pi - \arccos\left(\frac{\tilde{x}_i^T\tilde{x}_j}{d+1}\right)}{2\pi}. \label{eq:defp}
\end{align}

\begin{lemma}
Suppose at the end of iteration $k$, the prediction function is $f(x,w(k))$ for a new data point sampled i.i.d from the distribution $\mathcal{X}$. 
Then 
$$
\E_x(f(x,w(k)) - f_{KR}(x))^2 \le O\left(\frac{1}{\sqrt{n}}\right) +\left(1 - \frac{c\eta}{2} \right)^{k}(Y^TH^{-1}Y+O(1)) \text{ w.p. } 1-\delta
$$
when $\kappa^2 = O\left(\frac{c\delta}{d^2n^{1.5}}\right)$, $\eta = O\left(\frac{c}{(dn)^2}\right), m \ge poly(d,n,C_y,1/c,1/\delta).$
\label{lem:gen}
\end{lemma}
\textbf{Outline of the Proof: } We will prove this lemma using various concentration inequalities. The first step is to show that $f(x,w(k))$ is close to its linear approximation $f_L(x,w(k)) = \nabla^{\top} f(x,w(0))w(k)$ around $w(0).$ The second step is to show that $f_L(x,w(k))$ is close to the linear prediction function at $w_L^*,$ $f_L(x,w_L^*)$. The final step is to show that $f_L(x,w_L^*)$ is close to $f_{KR}(x)$ which close to the limit of $f_L(x,w_L^*)$ as $m\rightarrow\infty$. 

The characterization in Lemma \ref{lem:gen} is essential in showing the generalization result in Theorem \ref{th:gen} below.  The kernel $K(x_1,x_2)$ can be expressed as the inner product of infinite-dimensional feature vectors $\phi(x_1)$ and $\phi(x_2)$. Such a feature vector $\phi(.)$ exists because the Kernel is positive definite and symmetric \cite{mohri}. Indeed its easy to characterize the feature vector for $K(.,.)$. It is computed in Lemma \ref{lem:feature} below.

\begin{lemma}
\label{lem:feature}
Define a feature vector  $\phi(x)$ of data point $x$ as 
$$
\phi(x) := [\sqrt{d_0},\sqrt{d_1}x, \sqrt{d_2}x^{\otimes 2}, \sqrt{d_3}x^{\otimes 3},\ldots]\;\;
$$
where $x^{\otimes k}$ denotes the $k$ time Kronecker product of vector $x$ with itself and 
\begin{align*}
    & d_0 = 0.25 + \sum_{p\ge 1} \frac{c_{2p}}{(d+1)^{2p}},\;\;d_1 = 0.25 + \sum_{p\ge 1} \frac{2p c_{2p}}{(d+1)^{2p}}\;\;\\
    &d_k = \sum_{p\ge \lceil{k/2}\rceil} \frac{c_{2p}}{(d+1)^{2p}}{{2p}\choose {k}} \text{ for } k\ge 2,\;\;c_{2p} := \frac{(2p-3)!!(d+1)}{2\pi(2p-2)!!(2p-1)}.
\end{align*}

Then $K(x,y) = \phi(x)^T\phi(y)=\sum_{k\ge 0}d_k (x^{\top}y)^{k} = \tilde{x}_i^T\tilde{x}_j\frac{\pi - \arccos\left(\frac{\tilde{x}_i^T\tilde{x}_j}{d+1}\right)}{2\pi}$ for $\tilde{x} = \{x,1\} .$
\end{lemma}

In Theorem \ref{th:gen} below, we show that all functions $y=\phi^T(x)\bar{w},\|\bar{w}\|<\infty$ which are linear in the feature vector $\phi(x)$ are learnable by a finite width shallow ReLU network. Corollary \ref{cor:gen} provides some example functions in this class. This function class can be shown to be learnable from the generalization result in  \cite{arora2019fine}.
Their results are presented without using a bias term in each neuron, in which case, the infinite-width NTK RKHS is not a universal approximator. However, it is straightforward to extend their results to allow the addition of a bias term in each neuron, in which case, it is known that the infinite-width NTK RKHS is a universal approximator \cite{ji2019neural}. In the latter, more general case, their results match the results in this paper.
Our contribution is to provide an alternative proof of the result in \cite{arora2019fine}.  
\begin{theorem}
Suppose $y=\phi^{\top}(x)\bar{w}$ and $x$ is sampled from distribution $\mathcal{X}.$ Then there exists a constant $C>0$ such that 
$$
E_x(y-f(x,w(k)))^2\le C|\bar{w}|^2\sqrt{\frac{d\log(1/\delta)}{n}} \text{ w.p. greater than } 1-\delta \text{ over initialization}
$$\label{th:gen}
when $\kappa^2 = O\left(\frac{c\delta}{d^2n^{1.5}}\right)$, $\eta = O\left(\frac{c}{(dn)^2}\right), m \ge poly(d,n,C_y,1/c,1/\delta)$ and the number of gradient descent iterations $k=\Omega\left(\frac{\log\left(n(\|\bar{w}\|^2+1)\right)}{\eta c}\right).$
\end{theorem}
\textbf{Proof Sketch:} The argument follows from the proof of the noiseless case in \cite{bartlett2020benign}. Denote matrix $\Phi = [\phi(x_1) , \phi(x_2),\ldots \phi(x_n)].$ Together with Lemma \ref{lem:gen} we can show that $\E_x(y-f(x,w))^2\approx \E_x(y-f_{KR}(x))^2=\bar{w}^{T}P_{\phi}^{\perp}\E_x\phi(x)\phi^T(x)P_{\phi}^{\perp}\bar{w}$ where $P_{\phi}^{\perp}$ denotes the projection matrix to the null space of matrix $\Phi.$ Observe that the sample covariance matrix $n^{-1}\sum_{i=1}^n\phi(x_i)\phi^T(x_i)$ is orthogonal to the column space of $\Phi.$ This reduces the problem of bounding $\E_x(y-f_{KR}(x))^2$ to one of bounding the error between the population and sample covariance of $\phi(x)$. The result then follows from use of McDiarmid's inequality.

\begin{corollary}
If $y=(x^{\top}\beta)^{p} = \phi^{\top}(x) \bar{w}$ for $p=0,1,2,\ldots$ and $\|\beta\|\le 1$, then $$\bar{w} = [0,\ldots,0,\frac{1}{\sqrt{d_{p}}}\beta^{\otimes p},0,\ldots].$$ Therefore, by Theorem \ref{th:gen}, $
\E_x(y-f(x,w(k)))^2=O\left( (d+1)^p(p+1)^{1.5}\sqrt{\frac{\log(1/\delta)}{n}}\right) \text{ w.p. more than } 1-\delta $ for $p\ge 2$ and $\E_x(y-f(x,w(k)))^2=O\left(\sqrt{\frac{\log(1/\delta)}{n}}\right) \text{ w.p. more than } 1-\delta $ for $p=0,1$ when $\kappa^2 = O\left(\frac{c\delta}{d^2n^{1.5}}\right)$, $\eta = O\left(\frac{c}{(dn)^2}\right), m \ge poly(n,d^{p/2},1/c,1/\delta)$ and the number of gradient descent iterations $k=\Omega\left(\frac{\log\left(n\right)+p\log(d+1)}{\eta c}\right).$ Using the Stone–Weierstrass Theorem we can show that the set of polynomial functions are dense in the space of continuous functions in the compact input space and hence, the result here can be extended to include all continuous functions with appropriate approximation error.
\label{cor:gen}
\end{corollary}}
\section{Experiments}
\begin{figure}[h]
    \centering
    \subfigure[$\|P_0^{\perp}(w-w_L^*)\|^2$]{\label{fig:w_perp}\includegraphics[width=0.4\textwidth]{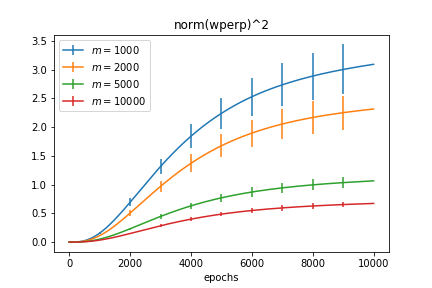}}
    \subfigure[$\|w-w_L^*\|^2$]{\includegraphics[width=0.4\textwidth]{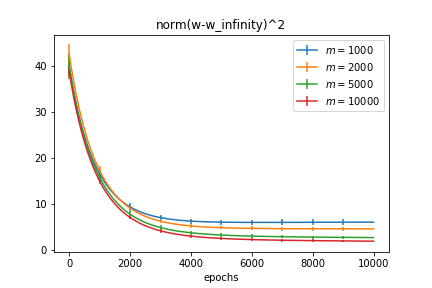}\label{fig:w_wsoln}}
    \subfigure[$\|w-w(0)\|^2$]{\includegraphics[width=0.4\textwidth]{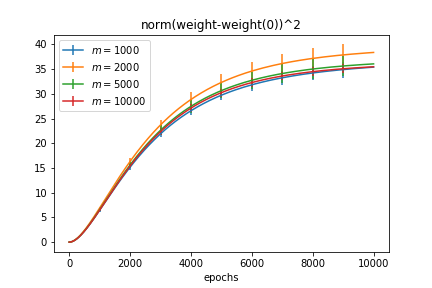}\label{fig:w_w0}}
    \caption{Gradient descent iterations for different widths of the network} \label{fig:example}
\end{figure}

We show an example with a synthetic dataset below. 
We generate $100$ data points from a  uniform distribution $[-1,1]^{5}$ in $\mathbb{R}^5$ and normalize it to have unit norm. The output is $y=(x^T\beta)^2$, where $x$ is the input and $\beta$ is chosen from a  uniform distribution $[-1,1]^{5}$ in $\mathbb{R}^5$. The output points are normalized by subtracting the empirical mean and then dividing by the empirical standard deviation.
We fit the data points using a shallow neural network of widths $m=1000,2000,5000,10000$ and mean squared loss. We perform full gradient descent with a learning rate of $0.01$ and do not train the last layer (i.e., the $a_i$) to be consistent with the theory presented in this paper. The experiment is repeated $5$ times with different initialization and the standard deviation is shown in Figure \ref{fig:example} $(b),(c)$. 
In Figure \ref{fig:example} $(b)$, we plot the different $\|w(t)-w_L^*\|^2$ for different widths across the iterations of gradient descent. Figure \ref{fig:example} $(a)$ shows that $\|P_0^{\perp}(w(t)-w_L^*)\|^2$ is upper bounded and the upper bound decreases with increase in width of the network.
Figure \ref{fig:example} $(c)$ plots distance of the iterates from initialization, $\|w(t)-w(0)\|^2$. We can see that the there is no clear trend on the bound for $\|w(t)-w(0)\|^2$ with increase in $m$ whereas $\|w(t)-w_L^*\|^2$ decreases when $m$ increases.

\section{Conclusions}

It has been well known for a while that, contrary to traditional wisdom, overparameterized neural networks perform well in training and generalization performance. One intuition behind this is that gradient descent for linear regression performs implicit regularization, i.e., minimizes the network parameter vector among all parameter vectors which provides zero training loss. Motivated by this intuition, recent works have shown that gradient descent, with good initialization, carried out on loss functions of neural networks have good generalization properties. We examine this phenomenon more closely and show that gradient descent with appropriate initialization converges to a point very close to a minimum solution of a linear regression approximation to the neural network training problem. We further use this result to provide generalization guarantees on the prediction function output by gradient descent in the noise free case.


\acks{Research supported by Navy Grant N00014-19-1-2566, NSF Grants CPS ECCS 1739189, CCF 1934986, NSF/USDA Grant AG 2018-67007-28379 and ARO Grant W911NF-19-1-0379.}

\newpage
\bibliography{ref}
\section{Proof of Theorem \ref{thm:cont}}
\label{app:thmcont}
The proof of this theorem  is divided into section \ref{sec:vperp}, which provides a bound on $V_{\perp}$, section \ref{sec:vpar}, which bounds $V_{\parallel}$ and section \ref{sec:w-w_infty} which provides a bound on $\|w-w_L^*\|$ and further proves that that $w(t)$ converges by proving that it forms a Cauchy sequence. Section \ref{sec:probconditions} provides the probability conditions that are assumed to be true in section \ref{sec:vperp}-\ref{sec:w-w_infty}. 
\subsection{Probability conditions}
\label{sec:probconditions}
This lemma below is state proved in \cite{app}, it requires use of various concentration inequalities.

\begin{lemma}
\label{lemma:prob}
With probability greater than $1 - \delta$ over initialization $a,w(0)$ the following conditions are true.  Lemma \ref{lem:trainloss} is true and 
\begin{enumerate}
\item \label{ass:4} Suppose $S_w := \{w:\max_k |w_k - w_k(0)|\le R\;\;\forall k\in[m]\}$. Then,
\begin{align*}
   \sup_{w\in S_w} \|\nabla f - \nabla f_0\|^2_F &\le \frac{8dnR}{\sqrt{2\pi}\kappa} + 4\sqrt{2}dn\sqrt{\frac{\log\left(\frac{20}{\delta}\right)}{m}}
\end{align*}
\begin{align*}
    &\sup_{w\in S_w}\|\nabla^T f\nabla f - \nabla^T f_0\nabla f_0\|\le \frac{8dn^2R}{\kappa\sqrt{2\pi}} + 4\sqrt{2}dn^2\sqrt{\frac{\log\left(\frac{20}{\delta}\right)}{m}}
\end{align*}

\item \label{ass:f0} At initialization $\|f_0\|$ is bounded, $\|f_0\| \le \frac{10\kappa\sqrt{dn}}{\delta}.$
\item The perturbation between $\nabla^T f_0\nabla f_0$ and $\E_{w(0)} [\nabla^T f_0\nabla f_0]$ is bounded from above by
$$\|\nabla^T f_0\nabla f_0 - H\| \le \sqrt{\frac{32n^2d}{m}\log\left(\frac{20n}{\delta}\right)}.$$
$
$ 

\end{enumerate}
\end{lemma}

\subsection{Bound on $V_{\perp}$}
\label{sec:vperp}
\begin{align*}
    \dot{V}_{\perp} &= -\langle P_0^{\perp}(w - w_L^*),P_0^{\perp} \nabla f(f-Y)\rangle
\end{align*}
Since $P_0^{\perp}\nabla f_0 =0$
\begin{align*}
    \dot{V}_{\perp} &= -\langle P_0^{\perp}(w - w_L^*),P_0^{\perp} (\nabla f - \nabla f_0) (f-Y)\rangle
\end{align*}
Applying the Cauchy-Schwarz inequality,
\begin{align*}
    \dot{V}_{\perp} &\le   \sqrt{V_{\perp}}\|\nabla f - \nabla f_0\| \|f-Y\|.
\end{align*}
From Lemma \ref{lemma:prob} $\|\nabla f - \nabla f_0\|= O\left(\frac{dn\sqrt{C_y}}{\sqrt{c\delta}m^{0.25}}\right)$ and from Lemma \ref{lem:trainloss} $\|f-Y\|^2\le \exp\left(- ct/2\right) \|f_0-Y\|^2.$  So
\begin{align*}
    \sqrt{V_{\perp}} = \int_0^{t}  \dot{\sqrt{V_{\perp}}}ds \le O\left(\frac{dn\sqrt{C_y}}{\sqrt{c\delta}m^{0.25}}\right) \int_0^{\infty}  \exp\left(- cs/4\right) \|f_0-Y\|ds = O\left(\frac{(dnC_y)^{1.5}}{c^{1.5}\delta^{1.5} m^{0.25}}\right).
\end{align*}

\subsection{Bound on $V_{\parallel}$}
\label{sec:vpar}
We can decompose $\dot{w}$ as
\begin{align*}
    \dot{w} = -\nabla f_0 \nabla^T f_0 (w-w_L^*)  +  (\nabla f_0 - \nabla f)(f-Y) + \nabla f_0(\nabla^T f_0w-f).
\end{align*}
Define
\begin{align*}
    e_w &:= (\nabla f_0 - \nabla f)(f-Y) + \nabla f_0(\nabla^T f_0w-f).
\end{align*}
Plugging this into $\dot{V}_{\parallel},$ we get
\begin{align}
    \dot{V}_{\parallel} = -\langle P_0(w - w_L^*), \nabla f_0 \nabla^T f_0 (w-w_L^*)\rangle +  \langle P_0(w - w_L^*), P_0e_w\rangle\label{eq:vapr}
\end{align}
First, we are going to bound each term in $\|e_w\|$ below.
\begin{itemize}
    \item $\|(\nabla f_0 - \nabla f)(f-Y)\|\le O\left(\frac{dn\sqrt{C_y}}{\sqrt{c\delta}m^{0.25}}\right) \exp\left(- ct/4\right) \|f_0-Y\|= O\left(\frac{(dnC_y)^{1.5}}{\sqrt{c}\delta^{1.5} m^{0.25}}\right)$
    \item Using $\nabla^T f_0w-f = \int_0^t (\nabla^T f_0\dot{w}-\dot{f}) ds $ and $\dot{f} = \nabla^T f\dot{w}$ 
    \begin{align*}
        \|\nabla^T f_0w-f\|  &\le \int_0^{\infty} \|(\nabla^T f_0-\nabla^T f)\dot{w}\| dt \le \int_0^{\infty} \|(\nabla^T f_0-\nabla^T f)\| \|\nabla f \| \|f-Y\| dt\\ 
        &\le \int_0^{\infty} \|(\nabla f_0-\nabla f)\| \|\nabla f \| \exp\left(- ct/4\right) \|f_0-Y\| dt  = O\left( \frac{(dn)^2C_y^{1.5}}{c^{1.5}\delta^{1.5}m^{0.25}}\right)
    \end{align*}
    
\end{itemize}

Hence $\|e_w\|=O\left(\frac{(dn)^{2.5}C_y^{1.5}}{c^{1.5}\delta^{1.5}m^{0.25}}\right)$. Using this bound on $\|e_w\|$ together with Cauchy Schwarz inequality we can simplify \eqref{eq:vapr} as

\begin{align*}
    \dot{V}_{\parallel} \le  -\langle P_0(w - w_L^*), \nabla f_0 \nabla^T f_0 (w-w_L^*)\rangle +  \sqrt{V_{\parallel}}O\left(\frac{(dn)^{2.5}C_y^{1.5}}{c^{1.5}\delta^{1.5}m^{0.25}}\right)
\end{align*}
From Lemma \ref{lemma:prob} $\nabla^T f_0 \nabla f_0\succeq c/2$ when $m=O(n^2d\log(n/\delta)/c^2)$. Hence $\nabla f_0 \nabla^T f_0\succeq c/2 P_0$ . So
\begin{align*}
    \dot{V}_{\parallel} \le  -c/2V_{\parallel} +  \sqrt{V_{\parallel}}O\left(\frac{(dn)^{2.5}C_y^{1.5}}{c^{1.5}\delta^{1.5}m^{0.25}}\right).
\end{align*}
Solving the above differential equation implies
\begin{align*}
    \sqrt{V_{\parallel}}\le \exp(-c/2t) \sqrt{V_{\parallel}(0)} +O\left(\frac{(dn)^{2.5}C_y^{1.5}}{c^{2.5}\delta^{1.5}m^{0.25}}\right).
\end{align*}

\subsection{Convergence of $w$}

In this subsection, we first show that the results in the previous two subsections imply that $\|w-w_L^*\|$ is bounded.
\label{sec:w-w_infty}
Combining results on $V_{\perp}$ and $V_{\parallel}$,
\begin{align*}
    \|w - w_L^*\| \le \sqrt{V_{\perp}} + \sqrt{V_{\parallel}} \le \exp(-c/2t) \sqrt{V_{\parallel}(0)} + O\left(\frac{(dn)^{2.5}C_y^{1.5}}{c^{2.5}\delta^{1.5}m^{0.25}}\right)
\end{align*}

\label{sec:convergescont}
We know that
$$w(t) = w(0) + \int_0^t \nabla f (f-Y) dt.$$
For any $t>s>T,$ we show that $\|w(t) - w(s)\| = O(\exp(-T)).$ {{Then it will follow that $\{w(t_n)\}_{n\ge 1}$ is a Cauchy sequence for any sequence $t_1\le t_2\ldots$ and that it converges.}}

\begin{align*}
    \|w(t) - w(s)\| & \le \int_s^t\|\nabla f (f-Y)\|dt \le \int_s^t \sqrt{nd}\exp\left(-\frac{ct}{4}\right)\|(f_0-Y)\|dt = O\left(\frac{dnC_y}{c\delta}\right) \exp(-T)
\end{align*}
 This shows that $w(t)\rightarrow w^*$ as $t\rightarrow\infty$ for some $w^*.$

\section{Proof of Lemma \ref{lem:psd}}
We can write the $H_{ij} =  \frac{\tilde{x}_i^{\top}\tilde{x}_j}{4}+\sum_{p\ge 1} c_{2p} (\frac{\tilde{x}_i^{\top}\tilde{x}_j}{d+1})^{2p},c_{2p} = \frac{(2p-3)!!(d+1)}{2\pi(2p-2)!!(2p-1)}.$ Consider the sequence of $n\times n$ matrices $H^{(p)},p\ge 1$ where we define the $i,j$ element of $H^{(p)}$ as $H^{(p)}_{ij} =  (\frac{\tilde{x}_i^{\top}\tilde{x}_j}{d+1})^{2p}.$ We can write $H = 0.25\tilde{X}^T\tilde{X}+\sum_{p\ge1}c_{2p}H^{(p)}$ where $\tilde{X}=[\tilde{x}_1\ldots \tilde{x}_n].$ Note that $H^{(p)},p\ge 1$ are positive semidefinite as we can write it as $H^{(p)} = \frac{1}{(d+1)^{2p}} \tilde{X}^{(p)T}\tilde{X}^{(p)},\tilde{X}^{(p)} = [\tilde{x}_1^{\otimes 2p}\ldots \tilde{x}_n^{\otimes 2p}].$ So $H$ is a positive semidefinite matrix. We will lower bound the smallest eigenvalue of $H$ by computing the smallest eigenvalue for each of $H^{(p)}$ using Gershgorin circle theorem. 
\begin{align*}
    \lambda_{\min}(H) = \min_{u:\|u\|=1} u^THu \ge\sum_{p\ge 1}c_{2p}\min_{u:\|u\|=1}u^TH^{(p)}u  = \sum_{p\ge 1}c_{2p}\lambda_{\min}(H^{(p)})
\end{align*}
The diagonal elements of $H^{(p)}$ are $1.$ Hence by Gershgorin circle theorem the eigenvalues of $H^{(p)}$ can be lower bounded by $1 - \max_i\sum_{j:j\neq i}H^{(p)}_{ij}.$ This bound is positive when $\max_i\sum_{j:j\neq i}H^{(p)}_{ij}$ is less than $1.$ Denote $\cos\theta_{\min} := \max_{i\neq j}\frac{\tilde{x}_j^T\tilde{x}_j}{d+1}.$ So $\max_i\sum_{j:j\neq i}H^{(p)}_{ij} \le (n-1)(\cos\theta_{\min})^{2p}.$ For all $p\ge k=\frac{\log(2n)}{\log(1/\cos\theta_{\min})},$   $(n-1)(\cos\theta_{\min})^{2p}\le 0.5.$ Hence 
\begin{align*}
    &\lambda_{\min}(H) \ge 0.5\sum_{p\ge k}c_{2p} \ge\sum_{p\ge k} \frac{d+1}{8\pi\sqrt{p-1}(2p-1)}\ge \int_{k+1}^{\infty} \frac{d+1}{8\pi\sqrt{x-1}(2x-1)}dx\ge \frac{d+1}{8\pi\sqrt{k+1}}\\
    &= \frac{d+1}{8\pi}\left(\sqrt{\frac{\log(1/\cos\theta_{\min})}{\log(2n/\cos\theta_{\min})}}\right)
\end{align*}
where we use Wallis' inequality to lower bound the ratio of the double factorial \cite{chen2005best}.

We can also get a simple upper bound on the minimum eigenvalue of $\lambda_{\min}(H).$ Suppose $a,b=\arg\max_{ij}x_i^Tx_j$. Choose vector $u\in\mathbb{R}^n$ with $u_a=\frac{1}{\sqrt{2}},u_b = \frac{1}{\sqrt{2}}.$
\begin{align*}
    \lambda_{\min}(H) = \min_{u:\|u\|=1} u^THu \le \frac{1}{2} (H_{aa}+H_{bb}-2H_{ab})=0.5(d+1)\left(1-\left(1-\frac{\theta_{\min}}{\pi}\right)\cos\theta_{\min}\right)
\end{align*}
Using the identity $1-\theta^2/2\le \cos\theta\le 1-\theta^2/4,\theta<1 $ we arrive at the approximations.
\section{Proof of Theorem \ref{th:GD}}
The fact that gradient descent achieves zero training in overparametrized networks is proved in \cite{du18}. 

\begin{lemma}[\cite{du18}]
The discrete time gradient descent algorithm achieves zero loss if $m= \Omega\left(\frac{d^4C_y^2n^6}{c^4\delta^3}\right),\kappa=1$ and $\eta = O\left(\frac{c}{d^2n^2}\right).$ 
With probability more than $1-\delta$ over initialization, $ \|f_k-Y\|^2\le \left(1-\frac{c\eta}{2}\right)^k \|f_0-Y\|^2$. Also, the weights stay close to initialization, $\|w_k(k)-w_k(0)\|= O\left(\frac{dnC_y}{\delta c\sqrt{m}}\right)$ for $k=\{0,1,\ldots\}.$
\label{du:gd}
\end{lemma}
\subsubsection{Bound on $V_{\perp}$}
\label{sec:vperpdiscrete}
Since, $P_0^{\perp}(w(0) - w_L^*) = 0,$ we can use a telescoping expansion of $P_{0}^{\perp}(w(k+1) - w_L^*)$ followed by the use of triangle inequality to get
\begin{align*}
    \sqrt{V_{\perp}(k+1)} &\le \sum_{l=0}^{k} \|P_0^{\perp}(w(l+1) - w(l))\|\\
    &= \eta \sum_{l=0}^{k}\|P_0^{\perp}(\nabla f_l - \nabla f_0) (f_l-Y) \|\\
    &\le \eta \sum_{l=0}^{k}O\left(\frac{dn\sqrt{C_y}}{\sqrt{c\delta}m^{0.25}}\right) \|f_l - Y\|\\
    &\le \eta O\left(\frac{dn\sqrt{C_y}}{\sqrt{c\delta}m^{0.25}}\right) \sum_{l=0}^{k}(1-c\eta)^{l/2}\|f_0 - Y\|\\
    &\le \eta O\left(\frac{dn\sqrt{C_y}}{\sqrt{c\delta}m^{0.25}}\right) \frac{1}{1 - \sqrt{1-c\eta}}\|f_0 - Y\|\le O\left(\frac{ dn\sqrt{C_y}}{c\sqrt{c\delta}m^{0.25}}\right)\|f_0-Y\|
\end{align*}

\subsubsection{Bound on $V_{\parallel}$}
\label{vparalleldiscrete}
We have
\begin{align}
    V_{\parallel}(k+1) =& \|P_0(w_{k+1} - w_L^*)\|^2 = \|P_0(w_{k} - w_L^*)\|^2 + \|P_0(w_{k+1} - w_k)\|^2\nonumber\\
    &+ 2\langle P_0(w_{k} - w_L^*), P_0(w_{k+1} - w_k)\rangle \label{eq:Vp}
\end{align}

We can expand $P_0(w_{k+1} - w_k)$ as
\begin{align}
    P_0(w_{k+1} - w_k) &= -\eta P_0\nabla f_k(f_k-Y)\nonumber\\
    & = -\eta(P_0(\nabla f_k - \nabla f_0)(f_k-Y) + \nabla f_0(f_k-\nabla ^Tf_0w_k) + \nabla f_0\nabla^T f_0(w_k - w_L^*))\nonumber\\
    &= e_w -\eta \nabla f_0\nabla^T f_0(w_k - w_L^*)) \label{eq:p0}
\end{align}
where we define $e_w := -\eta(P_0(\nabla f_k - \nabla f_0)(f_k-Y) + \nabla f_0(f_k-\nabla ^Tf_0w_k)).$
We can analyse the terms in \eqref{eq:Vp} separately as below using the decomposition in \eqref{eq:p0}.
\begin{itemize}
    \item $\|P_0(w_{k+1} - w_k)\|^2$:
    Using the bound on $\|\nabla f - \nabla f_0\|$ from Lemma \ref{lemma:prob},
    \begin{align*}
        \|P_0(w_{k+1} - w_k)\| &\le \eta O\left(\frac{dn\sqrt{C_y}}{\sqrt{c\delta}m^{0.25}}\right) \|f_k-Y\| + \eta\sqrt{dn}\|f_k - \nabla^T f_0 w_k\| + \eta dn  \|P_0(w_k - w_L^*)\|  
    \end{align*}
    \begin{align*}
        f_k - \nabla^T f_0 w_k &= \sum_{l=1}^k f_l - \nabla^T f_0 w_l - f_{l-1} - \nabla^T f_0 w_{l-1}  = \sum_{l=1}^k (\nabla f - \nabla f_0)^T(w_l - w_{l-1})\\
        &= \eta \sum_{l=1}^k (\nabla f - \nabla f_0)^T\nabla f_{l-1}(f_{l-1} - Y)
    \end{align*}
    By using the triangle inequality,
    \begin{align*}
        \|f_k - \nabla^T f_0 w_k\| &\le \eta \sum_{l=1}^k O(\frac{(dn)^{1.5}\sqrt{C_y}}{\sqrt{c\delta}m^{0.25}})\|f_{l-1} - Y\|\le \eta \sum_{l=1}^k O(\frac{(dn)^{1.5}\sqrt{C_y}}{\sqrt{c\delta}m^{0.25}})(1 - c\eta)^{l/2}\|f_{0} - Y\|\\
        &\le   O\left(\frac{(dn)^{1.5}\sqrt{C_y}}{c\sqrt{c\delta}m^{0.25}}\right)\|f_{0} - Y\|\\
        &= O\left(\frac{(dn)^{2}C_y^{1.5}}{c^{1.5}\delta^{1.5} m^{0.25}}\right)
    \end{align*}
    Using this bound we can upper bound $\|P_0(w_{k+1} - w_k)\|$ as
    \begin{align}
        \|P_0(w_{k+1} - w_k)\| &\le \eta O\left(\frac{dn\sqrt{C_y}}{\sqrt{c\delta}m^{0.25}}\right) (1-c\eta)^{k/2}\|f_0 - Y\| + \eta O\left(\frac{(dn)^{2.5}C_y^{1.5}}{c^{1.5}\delta^{1.5} m^{0.25}}\right) + \eta dn \sqrt{V_{\parallel}}\nonumber\\
        &\le \eta O\left(\frac{(dn)^{2.5}C_y^{1.5}}{c^{1.5}\delta^{1.5} m^{0.25}}\right) + \eta dn \sqrt{V_{\parallel}}
    \end{align}
    So $\|P_0(w_{k+1} - w_k)\|^2$ can be bounded as
    \begin{align}
        \|P_0(w_{k+1} - w_k)\|^2 &\le O\left(\frac{\eta^2(dn)^{5}C_y^{3}}{c^3\delta^3m^{0.5}}\right) + \eta^2 (dn)^2 V_{\parallel} + O\left(\frac{\eta^2(dn)^{3.5}C_y^{1.5}}{c^{1.5}\delta^{1.5} m^{0.25}}\right)  \sqrt{V_{\parallel}}\nonumber.
    \end{align}
    Using the bound $\sqrt{V_{\parallel}}\le 1 + V_{\parallel}$
    \begin{align}
        \|P_0(w_{k+1} - w_k)\|^2 &\le O\left(\frac{\eta^2(dn)^{5}C_y^{3}}{c^3\delta^3m^{0.5}}\right) + \eta^2 (dn)^2 V_{\parallel} + O\left(\frac{\eta^2(dn)^{3.5}C_y^{1.5}}{c^{1.5}\delta^{1.5} m^{0.25}}\right) + O\left(\frac{\eta^2(dn)^{3.5}C_y^{1.5}}{c^{1.5}\delta^{1.5} m^{0.25}}\right)V_{\parallel}\\
        &\le O\left(\frac{\eta^2(dn)^{5}C_y^{3}}{c^3\delta^3m^{0.25}}\right) + \left(\eta^2 (dn)^2  + O\left(\eta^2\frac{(dn)^{3.5}C_y^{1.5}}{c^{1.5}\delta^{1.5} m^{0.25}}\right)\right)V_{\parallel} \label{eq:Vpapallel}
    \end{align}
    \item $\langle P_0(w_{k} - w_L^*), P_0(w_{k+1} - w_k)\rangle$: We can expand $P_0(w_{k+1} - w_k)$ as in  \eqref{eq:p0} and use Cauchy-Schwarz inequality to get
    \begin{align}
        \langle P_0(w_{k} - w_L^*), P_0(w_{k+1} - w_k)\rangle & \le \|e_w\|\sqrt{V_{\parallel}} - \eta \langle P_0(w_{k} - w_L^*), \nabla f_0\nabla^T f_0(w_k - w_L^*))\rangle\nonumber
    \end{align}
    Using the bound $\nabla f_0\nabla^Tf_0 \succeq c P_0$ when $m\ge n^2d\log(n/\delta)/c^2$ from Lemma \ref{lemma:prob}
    \begin{align}
        \langle P_0(w_{k} - w_L^*), P_0(w_{k+1} - w_k)\rangle & \le \|e_w\|\sqrt{V_{\parallel}} - c\eta V_{\parallel}.
    \end{align}
    Now we can use the bound $\|e_w\| = \eta O\left(\frac{(dn)^{2.5}C_y^{1.5}}{c^{1.5}\delta^{1.5} m^{0.25}}\right)$ and $\sqrt{V_{\parallel}}\le 1+V_{\parallel}$ to have
    \begin{align}
        \langle P_0(w_{k} - w_L^*), P_0(w_{k+1} - w_k)\rangle & \le - \left(c\eta-\eta O\left(\frac{(dn)^{2.5}C_y^{1.5}}{c^{1.5}\delta^{1.5} m^{0.25}}\right)\right)V_{\parallel}+\eta O\left(\frac{(dn)^{2.5}C_y^{1.5}}{c^{1.5}\delta^{1.5} m^{0.25}}\right).\label{eq:11}
    \end{align}
\end{itemize}
We are now ready to combine the bounds from \eqref{eq:Vpapallel} and \eqref{eq:11} into the recursion in \eqref{eq:Vp}.
    If we have $\eta\le \frac{c}{2(dn)^2},$ and $m=\Omega\left(\frac{(dn)^{10}C_y^6}{c^{10}\delta^6}\right)$
    \begin{align}
        V_{\parallel}(k+1) &\le V_{\parallel}(k) +  O\left(\frac{\eta^2(dn)^{5}C_y^{3}}{c^3\delta^3m^{0.25}}\right) +  \left(\eta^2 (dn)^2  + O\left(\eta^2\frac{(dn)^{3.5}C_y^{1.5}}{c^{1.5}\delta^{1.5} m^{0.25}}\right)\right)V_{\parallel}(k)\nonumber\\
        &- \left(c\eta-\eta O\left(\frac{(dn)^{2.5}C_y^{1.5}}{c^{1.5}\delta^{1.5} m^{0.25}}\right)\right)V_{\parallel}+\eta O\left(\frac{(dn)^{2.5}C_y^{1.5}}{c^{1.5}\delta^{1.5} m^{0.25}}\right)\\
        &\le \left(1 - \frac{c\eta}{2} \right)V_{\parallel}(k) + O\left(\frac{\eta (dnC_y)^{3}}{c^2\delta^3m^{0.25}}\right)\\
        &\le \left(1 - \frac{c\eta}{2} \right)^kV_{\parallel}(0) + O\left(\frac{ (dnC_y)^{3}}{c^3\delta^3m^{0.25}}\right).
    \end{align}
    where the final step follows from induction.
\subsubsection{Putting it all together}
\label{putittogether}
We combine the bounds on $V_{\parallel}$ and $V_{\perp}$ to get
\begin{align}
    \|w_k-w_L^*\| \le \sqrt{V_{\parallel}} + \sqrt{V_{\perp}} &\le O\left(\frac{ dn\sqrt{C_y}}{c\sqrt{c\delta}m^{0.25}}\right)\|f_0-Y\| + \sqrt{\left(1 - \frac{c\eta}{2} \right)^kV_{\parallel}(0) + O\left(\frac{ (dnC_y)^{3}}{c^3\delta^3m^{0.25}}\right)}\nonumber\\
    &\le\left(1 - \frac{c\eta}{2} \right)^{k/2}\|w_0-w_L^*\| + O\left(\frac{(dnC_y)^{1.5}}{c^{1.5}\delta^{1.5} m^{0.125}}\right)
\end{align}

\subsubsection{Convergence of iterates for GD}
\label{sec:convergence}
Similar to the continuous time dynamics of gradient flow, we show that the iterates $w_k$ form a Cauchy sequence. For $n>m>N,$
\begin{align*}
    \|w_n-w_m\| &\le \eta \sum_{k=m}^n \|\nabla f_k(f_k - Y)\| \le \eta\sqrt{dn} \sum_{k=m}^n (1-c\eta)^{k/2}\|(f_0 - Y)\|\\
    &\le \eta\sqrt{dn}  (1-c\eta)^{m/2}\frac{1}{1 - \sqrt{1-c\eta}}\|(f_0 - Y)\| \\
    &\le \frac{2\sqrt{dn}}{c}  (1-c\eta)^{N/2}\|(f_0 - Y)\|
\end{align*}
Hence, $\{w_k\}_{k=1}^{\infty}$ converges.

\section{Proof of Lemma \ref{lem:gen}}

We can decompose $f(x,w(k))-f_{KR}(x)$ as 
\begin{align*}
     (f(x,w(k))-f_L(x,w(k))) + (f_L(x,w(k))-f_L(x,w_L^*)) + (\nabla^{\top}f(x,w(0))P_0^{\perp}w(0)) +\\ (\nabla^{\top}f(x,w(0))\nabla f_0(\nabla^{\top}f_0\nabla f_0)^{-1}Y - f_{KR}(x)). 
\end{align*}
We bound each of the above terms below. In the derivation below we use the bounds resulting from Theorem \ref{th:GD}, Lemma \ref{du:gd}  and Lemma \ref{lemma:prob} deterministically. Subsection \ref{subsec:probprediction} shows the probabilistic events that are assumed to be true w.p. greater than $1-\delta$ in the proof below.

\begin{itemize}
    \item We begin by showing that $\E_x (f(x,w(k)) - f_L(x,w(k)))^2$ is small. We can expand $f(x,w(k)) - f_L(x,w(k))$ as
    \begin{align*}
        \frac{1}{\sqrt{m}}\sum_{j}a_j( \sigma(w_j^{\top}(k)\tilde{x}) - \sigma'(w_j^{\top}(0)\tilde{x})w_j^{\top}(k)\tilde{x} ).  
    \end{align*}
    Define $S(x,w(0)):=\{ j\in[m]: \sigma'(w_j^{\top}(0)\tilde{x})\neq \sigma'(w_j^{\top}(k)\tilde{x})\},$ the set of indices from $1\ldots m$ for which $w_j^{\top}(0)\tilde{x}$ has a different sign than $w_j^{\top}(k)\tilde{x}$. This implies that $f(x,w(k)) - f_L(x,w(k))$ can be equivalently written as
    \begin{align*}
        \frac{1}{\sqrt{m}}\sum_{j\in S(x,w(0))}a_j( \sigma(w_j^{\top}(k)\tilde{x}) - \sigma(w_j^{\top}(0)\tilde{x}) + \sigma(w_j^{\top}(0)\tilde{x}) -  \sigma'(w_j^{\top}(0)\tilde{x})w_j^{\top}(k)\tilde{x} ).
    \end{align*}
    We can use triangle inequality followed by 1-Lipschitz property of $\sigma(.)$ to bound $|f(x,w(k)) - f_L(x,w(k))|$ by
    \begin{align*}
        \frac{2}{\sqrt{m}}\sum_{j\in S(x,w(0))} |w_j^{\top}(k)\tilde{x} - w_j^{\top}(0)\tilde{x}|.
    \end{align*}
    Denote the bound on $\|w_j(k) - w_j(0)\|$
    from Lemma \ref{du:gd} as $R:= O\left(\frac{dnC_y}{\delta c\sqrt{m}}\right)$. So
    \begin{align}
        |f(x,w(k)) - f_L(x,w(k))|^2\le \frac{4dR^2|S(x,w(0))|^2}{m},\label{eq:fminusfl}
    \end{align}
    where $|S(x,w(0))|$ denotes the cardinality of set $S(x,w(0)).$ From section \ref{subsec:probprediction}, we have that
    \begin{align*}
        \E_x |S(x,w(0))|^2 \le \frac{\E_{x,w(0)} |S(x,w(0))|^2}{\delta}
    \end{align*}
    When $|w_j^{\top}(0)\tilde{x}|\le |w_j^{\top}(0)\tilde{x} - w_j^{\top}(k)\tilde{x}|$ there is a difference in sign between $w_j^{\top}(0)\tilde{x}$ and $w_j^{\top}(k)\tilde{x}$ . Hence
    \begin{align*}
        \E_{x,w(0)} |S(x,w(0))|^2 \le \E_{x,w(0)} \left(\sum_j \mathds{1}\{ |w_j^{\top}(0)\tilde{x}|\le |w_j^{\top}(0)\tilde{x} - w_j^{\top}(k)\tilde{x} | \}\right)^2.
    \end{align*}
    By triangle inequality followed by Cauchy-Schwartz
    \begin{align*}
        \E_{x,w(0)}|S(x,w(0))|^2 \le  \sum_{j,l} \sqrt{\E_{x,w_j(0)}\mathds{1}\{ |w_j^{\top}(0)\tilde{x}|\le R \}\E_{x,w_l(0)}\mathds{1}\{ |w_l^{\top}(0)\tilde{x}|\le R \}}.
    \end{align*}
    Since $w_j(0)$ is distributed $\mathcal{N}(0,\kappa^2I_d)$ and $x$ is a distributed as $\mathcal{X}$ which is supported on the ball $\|x\|=\sqrt{d},$ $w_j^{\top}(0)\tilde{x}$ is distributed as $\mathcal{N}(0,\kappa^2(d+1))$. Hence $\Prob( |w_j^{\top}(0)\tilde{x}|\le R )$ is bounded by $\frac{R}{\sqrt{2\pi d}\kappa}.$ This shows
    \begin{align*}
        \E_{x,w(0)}|S(x,w(0))|^2 = O\left( \frac{m^2R}{\sqrt{d}\kappa}\right) 
    \end{align*}
    Plugging this bound back into \eqref{eq:fminusfl} we have
    \begin{align*}
        E_x|f(x,w(k)) - f_L(x,w(k))|^2= O\left(\frac{\sqrt{d}mR^3}{\delta\kappa }\right) = O\left(\frac{d^{3.5}n^{3}C_y^3}{\kappa\delta^4 c^3m^{0.5}}\right)
    \end{align*}
    \item We can upper bound the term $E_x|f_L(x,w(k)) - f_L(x,w_L^*)|^2\le \|w(k)-w_L^*\|^2. $ Using the bound on $\|w(k)-w_L^*\|$ from Theorem \ref{th:GD}
    \begin{align*}
        E_x|f_L(x,w(k)) - f_L(x,w_L^*)|^2\le \left(1 - \frac{c\eta}{2} \right)^{k/2}\|w(0)-w_L^*\| + O\left(\frac{(dnC_y)^{1.5}}{c^{1.5}\delta^{1.5} m^{0.125}}\right)
    \end{align*}
    From the definition of $w_L^* $ we can write $w(0)-w_L^* = P_0w(0) -\nabla f_0 (\nabla^T f_0\nabla f_0)^{-1}Y.$ Hence
    \begin{align*}
        \|w(0)-w_L^*\|^2&\le 2\|P_0w(0)\|^2 +2\|\nabla f_0 (\nabla^T f_0\nabla f_0)^{-1}Y\|^2\\
        &= 2f_0^T(\nabla f_0^{\top}\nabla f_0)^{-1}f_0 +2Y^T(\nabla f_0^{\top}\nabla f_0)^{-1}Y
    \end{align*}
    Using the bound on $\|f_0\|^2$ from section \ref{subsec:probprediction} $f_0^T(\nabla f_0^{\top}\nabla f_0)^{-1}f_0= O\left(\frac{1}{\sqrt{dn}}\right) $  when $\kappa^2=O\left(\frac{c\delta}{(dn)^{1.5}}\right).$ Now consider the term $Y^T(\nabla^T f_0\nabla f_0)^{-1}Y.$ 
    \begin{align*}
        Y^T(\nabla^T f_0\nabla f_0)^{-1}Y= Y^TH^{-1}Y+Y^T((\nabla^T f_0\nabla f_0)^{-1}-H^{-1})Y. 
    \end{align*}
    Using the identity $A^{-1}-B^{-1} = A^{-1}(B - A)B^{-1}$ and the concentration results from Lemma \ref{lemma:prob}
    \begin{align*}
        Y^T((\nabla^T f_0\nabla f_0)^{-1}-H^{-1})Y \le \frac{n^2C_y^2\sqrt{d}}{c^2\sqrt{m}}\sqrt{\log\left(\frac{n}{\delta}\right)}= O\left(1\right)
    \end{align*}
    when $m=\Omega(dn^4C_y^4\log(n/\delta)/c^4)$.
    Thus $\|w(0)-w_L^*\|^2\le Y^TH^{-1}Y+O\left(1\right).$
    \item Consider $\nabla^{\top}f(x,w(0))P_0^{\perp}w(0) = f(x,w(0)) - \nabla^{\top}f(x,w(0))\nabla f_0(\nabla f_0^{\top}\nabla f_0)^{-1}f_0.$ From section \ref{subsec:probprediction} 
    \begin{align}
        \E_{x} f^2(x,w(0))\le \frac{\kappa^2d}{2\delta} \label{eq:f1}
    \end{align}
    For the second term, note that $E_x(\nabla^{\top}f(x,w(0))\nabla f_0(\nabla f_0^{\top}\nabla f_0)^{-1}f_0)^2\le df_0^T(\nabla f_0^{\top}\nabla f_0)^{-1}f_0$ since $\E_x\nabla f(x,w(0))\nabla^{\top}f(x,w(0))\preccurlyeq dI$ and so
    \begin{align*}
        E_x(\nabla^{\top}f(x,w(0))\nabla f_0(\nabla f_0^{\top}\nabla f_0)^{-1}f_0)^2\le \frac{d\|f_0\|^2}{c}
    \end{align*}
    when $m=\Omega(n^2d\log(n/\delta)/c^2).$
    Using the bound on $\|f_0\|^2$ from section \ref{subsec:probprediction} 
    \begin{align*}
        \frac{\|f_0\|^2}{c} =O\left( \frac{\kappa^2dn}{c\delta}\right) 
    \end{align*}
    Combining this with \eqref{eq:f1}
    \begin{align*}
        \E_x (\nabla^{\top}f(x,w(0))P_0^{\perp}w(0))^2 \le \frac{\kappa^2d^2n}{c\delta}=O\left(\frac{1}{\sqrt{n}}\right)
    \end{align*}
    when $\kappa^2=O\left(\frac{c\delta}{d^2n^{1.5}}\right).$
    \item Finally, consider the term $\nabla^{\top}f(x,w(0))\nabla f_0(\nabla^{\top}f_0\nabla f_0)^{-1}Y - f_{KR}(x).$ We can expand it as
    \begin{align*}
        \left(\nabla^{\top} f(x,w(0))\nabla f_0-h^{\top}\right)(\nabla^{\top}f_0\nabla f_0)^{-1}Y + h^{\top}\left((\nabla^{\top}f_0\nabla f_0)^{-1}-H^{-1}\right)Y.
    \end{align*}
    Applying triangle inequality we can upper bound $\E_x\left(\nabla^{\top}f(x,w(0))\nabla f_0(\nabla^{\top}f_0\nabla f_0)^{-1}Y - f_{KR}(x)\right)^2$ as
    \begin{align}
        E_x\left(\frac{nC_y}{c^2} \|\nabla^{\top} f_0\nabla f(x,w(0))-h\|^2\right)+ \frac{(dnC_y)^2}{16}\|(\nabla^{\top}f_0\nabla f_0)^{-1}-H^{-1}\|^2\label{eq:third1}
    \end{align}
    First consider the left term in \eqref{eq:third1}. From section \ref{subsec:probprediction} the left term in \eqref{eq:third1} can be bounded by
    \begin{align*}
         \frac{nC_y}{c^2\delta}\sum_{i}\E_{x,w(0),x_i} \left(\frac{1}{m}\sum_{j}\sigma'(w_j^{\top}(0)\tilde{x})\sigma'(w_j^{\top}(0)\tilde{x}_i)\tilde{x}^{\top}\tilde{x}_i - K(x,x_i)\right)^2
    \end{align*}
    Since $w_j^{\top}(0)\tilde{x}_i$ and $w_j^{\top}(0)\tilde{x}$ are statistically independently for all $j$ given $x,x_i,$ we have
    \begin{align}
        E_x\left(\frac{nC_y}{c^2} \|\nabla^{\top} f(x,w(0))\nabla f_0-h^{\top}\|^2\right)
        &\le \frac{2nC_y}{m c^2\delta}  \label{eq:thirdleft}
    \end{align}
    Now consider the second term in \eqref{eq:third1}. Using the identity $A^{-1}-B^{-1} = A^{-1}(B - A)B^{-1}$ and the concentration results from Lemma \ref{lemma:prob}
    \begin{align}
        \frac{(dnC_y)^2}{16}\|(\nabla^{\top}f_0\nabla f_0)^{-1}-H^{-1}\|^2\le \frac{d^2n^4C_y^2}{c^4m} \log\left(\frac{20n}{\delta}\right)
    \end{align}
    when $m=O(n^2d\log(n/\delta)/c^2)$.
    Combining the above bound with \eqref{eq:thirdleft}
    \begin{align*}
        \E_x\left(\nabla^{\top}f(x,w(0))\nabla f_0(\nabla^{\top}f_0\nabla f_0)^{-1}Y - f_{KR}(x)\right)^2\le \frac{d^2n^4C_y^2\log n}{c^4m\delta}
    \end{align*}
\end{itemize}
\subsection{Probability conditions for Lemma \ref{lem:gen}}
\label{subsec:probprediction}
We can use a union bound to show the below events occur simultaneously with probability at least $1-\delta$
\begin{itemize}
    \item Theorem \ref{th:GD}, Lemma \ref{du:gd}  and Lemma \ref{lemma:prob} are true.
    \item By Markov inequality, we have that
    \begin{align*}
        \E_x |S(x,w(0))|^2 \le \frac{10\E_{x,w(0)} |S(x,w(0))|^2}{\delta} \text{ w.p. greater than } 1-\delta/10 
    \end{align*}
    \item Using Markov inequality 
    \begin{align*}
        \E_{x} f^2(x,w(0))\le \frac{10d\kappa^2}{\delta} \text{ w.p. greater than} 1 - \delta/10 
    \end{align*}
    \item  By Markov inequality  \begin{align*}
        \E_x\left( \|\nabla^{\top} f(x,w(0))\nabla f_0-h^{\top}\|^2\right)\le \E_{x,w(0),x_i}\left(\frac{10}{\delta } \|\nabla^{\top} f(x,w(0))\nabla f_0-h^{\top}\|^2\right)
    \end{align*}
    with probability greater than  $1-\delta/10.$
    \item Again, we apply Markov inequality. 
    \begin{align*}
        \frac{\|f_0\|^2}{c} \le \frac{\E \|f_0\|^2}{c\delta}\le \frac{10\kappa^2dn}{c\delta} \text{ w.p. greater than } 1-\delta/10.
    \end{align*}
\end{itemize}

\section{Proof of Lemma \ref{lem:feature}}
Note that $K(x,y) = \tilde{x}^T\tilde{y}\frac{\pi - \arccos\left(\frac{\tilde{x}^T\tilde{y}}{d+1}\right)}{2\pi} =  \frac{\tilde{x}^{\top}\tilde{y}}{4}+\sum_{p\ge 1} c_{2p} (\frac{\tilde{x}^{\top}\tilde{y}}{d+1})^{2p},c_{2p} = \frac{(2p-3)!!(d+1)}{2\pi(2p-2)!!(2p-1)}.$ Using $\tilde{x} = \{x,1\}$
\begin{align*}
    K(x,y) &= \frac{1}{4} +  \frac{x^{\top}y}{4} +\sum_{p\ge 1} c_{2p} \left(\frac{x^{\top}y+1}{d+1}\right)^{2p}\\
    & = \frac{1}{4} +  \frac{x^{\top}y}{4} +\sum_{p\ge 1} \frac{c_{2p}}{(d+1)^{2p}} \sum_{k=0}^{2p} {{2p}\choose k}(x^{\top}y)^k\\
    & = \left(\frac{1}{4} + \sum_{p\ge 1} \frac{c_{2p}}{(d+1)^{2p}}\right)+ \left(\frac{1}{4} + \sum_{p\ge 1} \frac{2pc_{2p}}{(d+1)^{2p}} \right) x^{\top}y +\sum_{k\ge 2}\sum_{p\ge \lceil k/2\rceil} \frac{c_{2p}}{(d+1)^{2p}}  {{2p}\choose k}(x^{\top}y)^k
\end{align*}
Denoting the coefficient of $(x^{\top}y)^k$ by $d_k$
\begin{align}
    K(x,y) = \sum_{k\ge 0} d_k (x^{\top}y)^k = \sum_{k\ge 0} d_k (x^{\otimes k})^{\top}y^{\otimes k} = \phi^{\top}(x) \phi(y)
\end{align}
where $\phi(x) = [\sqrt{d_0},\sqrt{d_1}x, \sqrt{d_2}x^{\otimes 2}, \sqrt{d_3}x^{\otimes 3},\ldots].$

\section{Proof of Theorem \ref{th:gen}}

Denote matrix $\Phi = [\phi(x_1) , \phi(x_2),\ldots \phi(x_n)]$ with $n$ columns. Note that $$f_{KR}(x) = \phi^{\top}(x)\Phi(\Phi^{\top}\Phi)^{-1}\Phi^{\top} \bar{w} = \phi^{\top}(x)P_{\Phi} \bar{w},$$
where $P_{\Phi}$ is the projection matrix onto the columns space of $\Phi.$ First we center the random variable $\phi(x)$ around its expectation and denote $\tilde{\phi}(x) = \phi(x)-\E_x \phi(x).$ Since the columns space of $\Phi$ remains unchanged due to this transformation, $P_{\tilde{\Phi}}=P_{\Phi}.$
So
\begin{align*}
    &E_x(y-f_{KR}(x))^2 = E_x ( \tilde{\phi}^{\top}(x)(I-P_{\tilde{\Phi}})\bar{w} + \E_x\phi^{\top}(x)(I-P_{\tilde{\Phi}})\bar{w})^2\\
    &\le 2 \bar{w}^{\top} (I-P_{\tilde{\Phi}})\E_x \tilde{\phi}(x)\tilde{\phi}^{\top}(x)(I-P_{\tilde{\Phi}})\bar{w} + 2\bar{w}^{\top} (I-P_{\Phi})\E_x \phi(x)\E_x\phi^{\top}(x)(I-P_{\Phi})\bar{w} .
\end{align*}
Using $\|I-P_{\tilde{\Phi}}\|\le 1$ and $ P^{\perp}_{\Phi}\phi(x_i) = 0\;\forall i\in[n],$ we can further upper bound
\begin{align}
    E_x(y-f_{KR}(x))^2 & \le 2 \|\bar{w}\|^2\| \E_x \tilde{\phi}(x)\tilde{\phi}^{\top}(x) - n^{-1}\sum_{i} \tilde{\phi}(x_i)\tilde{\phi}^{\top}(x_i)\|+ 2 \|\bar{w}\|^2\| n^{-1}\sum_{i} \tilde{\phi}(x_i)\|^2\label{eq:conc}
\end{align}
Now we will apply McDiarmids inequality to the first term in \eqref{eq:conc}. Note that typically one would need to use more involved concentration inequality for sample covariance matrix (like \cite{koltchinskii2014concentration}). But our data points $\tilde{\phi}(x)$ are bounded and hence we can simply use McDiarmids inequality. 
If we change one of the $\tilde{\phi}(x_i)$ with its i.i.d. copy then we can apply triangle inequality to show that $\| \E_x \tilde{\phi}(x)\tilde{\phi}^{\top}(x) - n^{-1}\sum_{i} \tilde{\phi}(x_i)\tilde{\phi}^{\top}(x_i)\|$ changes by a maximum of $2(d+1)/n.$ Hence by  McDiarmids inequality
\begin{align*}
    \| \E_x \tilde{\phi}(x)\tilde{\phi}^{\top}(x) - n^{-1}\sum_{i} \tilde{\phi}(x_i)\tilde{\phi}^{\top}(x_i)\|\le \E_{x_i} \| \E_x \tilde{\phi}(x)\tilde{\phi}^{\top}(x) - n^{-1}\sum_{i} \tilde{\phi}(x_i)\tilde{\phi}^{\top}(x_i)\|\\
    + \sqrt{\frac{(d+1)\log(1/\delta)}{n}} \text{ w.p. } 1-\delta.
\end{align*}
We can now upper bound $\E_{x_i} \| \E_x \tilde{\phi}(x)\tilde{\phi}^{\top}(x) - n^{-1}\sum_{i} \tilde{\phi}(x_i)\tilde{\phi}^{\top}(x_i)\|$ as
\begin{align*}
    \E_{x_i} \| \E_x \tilde{\phi}(x)\tilde{\phi}^{\top}(x) - n^{-1}\sum_{i} \tilde{\phi}(x_i)\tilde{\phi}^{\top}(x_i)\|&\le\sqrt{\E_{x_i} \| \E_x \tilde{\phi}(x)\tilde{\phi}^{\top}(x) - n^{-1}\sum_{i} \tilde{\phi}(x_i)\tilde{\phi}^{\top}(x_i)\|_F^2}  \\
    &= \sqrt{\E_{x_i} n^{-2} \sum_{i}\|\E_x \tilde{\phi}(x)\tilde{\phi}^{\top}(x) -  \tilde{\phi}(x_i)\tilde{\phi}^{\top}(x_i)\|_F^2 }\\
    &\le \sqrt{\frac{2(d+1)}{n}}
\end{align*}
as $\|\tilde{\phi}(x)\|\le d+1.$ Hence 
\begin{align}
    \|\E_x \tilde{\phi}(x)\tilde{\phi}^{\top}(x) - n^{-1}\sum_{i} \tilde{\phi}(x_i)\tilde{\phi}^{\top}(x_i)\| =O\left(  \sqrt{\frac{(d+1)\log(1/\delta)}{n}}  \right)\text{ w.p. mote than } 1-\delta.\label{eq:term1}
\end{align}
We will apply McDiarmids inequality to the second term in \eqref{eq:conc} as well. Since $\| n^{-1}\sum_{i} \tilde{\phi}(x_i)\|$ changes by a maximum of $2(d+1)/n$ by changing one of the $\tilde{\phi}(x_i)$ with an i.i.d copy
\begin{align}
    \| n^{-1}\sum_{i} \tilde{\phi}(x_i)\|\le \E_{x_i} \| n^{-1}\sum_{i} \tilde{\phi}(x_i)\| +\sqrt{\frac{(d+1)\log(1/\delta)}{n}} \text{ w.p. mote than } 1-\delta.\label{eq:term2}
\end{align}
The mean $\E_{x_i} \| n^{-1}\sum_{i} \tilde{\phi}(x_i)\|\le \sqrt{\E_{x_i} \| n^{-1}\sum_{i} \tilde{\phi}(x_i)\|^2}\le \sqrt{\frac{d+1}{n}}.$ Putting this together with \eqref{eq:term2}
\begin{align}
    \| n^{-1}\sum_{i} \tilde{\phi}(x_i)\|\le \sqrt{\frac{d+1}{n}} +\sqrt{\frac{(d+1)\log(1/\delta)}{n}} \text{ w.p. mote than } 1-\delta.\label{eq:term3}
\end{align}
Combining the result in  \eqref{eq:term3} and \eqref{eq:term1} with \eqref{eq:conc}, we arrive at the bound on $E_x(y-f_{KR}(x))^2$. Finally, using the bound on $E_x(y-f(x,w(k)))^2$ from Lemma \ref{lem:gen} we arrive at the result.

\section{Proof of Corollary 9}
If $y = \left(x^{\top}\beta\right)^p, \|\beta\|\le 1$ then $y$ can be equivalently written as 
\begin{align*}
    y = (\sqrt{d_p}x^{\otimes p})^{\top}\frac{1}{d_p}\beta^{\otimes p}  = \phi^{\top}(x)\bar{w}, \bar{w} = [0,\ldots,0,\frac{1}{\sqrt{d_{p}}}\beta^{\otimes p},0,\ldots].
\end{align*}

In order to apply Theorem \ref{th:gen} we need to compute an upper bound on $\|\bar{w}\|^2 = \frac{\|\beta\|^{2p}}{d_p}.$ It boils down to computing a lower bound on $d_p.$ $d_p\ge 0.25$ for $p = 0,1$. For $p\ge 2$    $d_p = \sum_{p'\ge \lceil{p/2}\rceil} \frac{c_{2p'}}{(d+1)^{2p'}}{{2p'}\choose {p}}.$ 
Using the results in \cite{chen2005best} to lower bound the ratio of double factorial,

$$c_{2p'} = \frac{(2p'-3)!!(d+1)}{2\pi(2p'-2)!!(2p'-1)}\ge\frac{d+1}{10(2p')^{1.5}}.$$
Combining this with the bound $ {{2p'}\choose {k}}\ge\left(\frac{2p'}{k}\right)^k$ 
$$\sum_{p'\ge \lceil{p/2}\rceil}\frac{c_{2p'}}{(d+1)^{2p'}}{{2p'}\choose {p}}\ge \sum_{p'\ge \lceil{p/2}\rceil} \frac{1}{10(2p')^{1.5}(d+1)^{2p'-1}}\left(\frac{2p'}{p}\right)^p\ge \frac{1}{10(p+1)^{1.5}(d+1)^{p}} $$
Hence $\|\bar{w}\|^2\le 10(p+1)^{1.5} (d+1)^p$ and we arrive at the result.

\end{document}